\documentclass[runningheads]{llncs}
\usepackage[T1]{fontenc}
\usepackage{pdfpages} % include pdfs
\usepackage{pgffor} % for loops

\makeatletter
\AtBeginDocument{\let\LS@rot\@undefined}
\makeatother

\def\supplementfilename{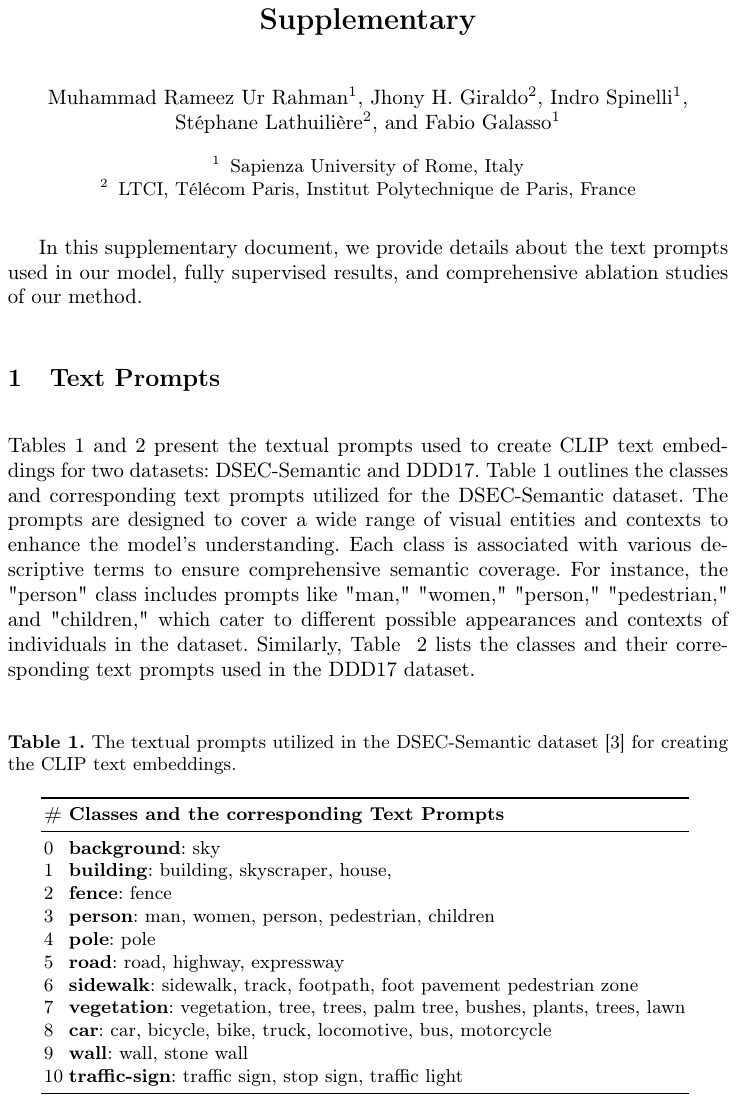}

\pdfximage{\supplementfilename}
\def\numbersupplementpages{\the\pdflastximagepages}

\newif\ifarXiv
\arXivtrue 

\usepackage{multicol}
\usepackage{multirow}
\usepackage{booktabs}
\usepackage{pifont}

\usepackage{xcolor}
\usepackage{graphicx}
\usepackage{maths}
\usepackage{arydshln}
\usepackage{hyperref}

\newcommand{\our}{OVOSE\xspace}
\usepackage{color,soul}
\begin{document}
\title{OVOSE: Open-Vocabulary Semantic Segmentation in Event-Based Cameras}

\author{Muhammad Rameez Ur Rahman\inst{1} \and
Jhony H. Giraldo\inst{2} \and
Indro Spinelli\inst{1} \and Stéphane Lathuilière\inst{2} \and Fabio Galasso\inst{1}}
\authorrunning{Rahman et al.}
\institute{Sapienza University of Rome, Italy\\
\email{\{rahman, spinelli, galasso\}@di.uniroma1.it}
\and
LTCI, Télécom Paris, Institut Polytechnique de Paris, France\\
\email{\{jhony.giraldo, stephane.lathuiliere\}@telecom-paris.fr}}
\maketitle              % typeset the header of the contribution
\begin{abstract}
Event cameras, known for low-latency operation and superior performance in challenging lighting conditions, are suitable for sensitive computer vision tasks such as semantic segmentation in autonomous driving.
However, challenges arise due to limited event-based data and the absence of large-scale segmentation benchmarks.
Current works are confined to closed-set semantic segmentation, limiting their adaptability to other applications.
In this paper, we introduce \our, the first Open-Vocabulary Semantic Segmentation algorithm for Event cameras. 
\our leverages synthetic event data and knowledge distillation from a pre-trained image-based foundation model to an event-based counterpart, effectively preserving spatial context and transferring open-vocabulary semantic segmentation capabilities.
We evaluate the performance of \our on two driving semantic segmentation datasets DDD17, and DSEC-Semantic, comparing it with existing conventional image open-vocabulary models adapted for event-based data.
Similarly, we compare \our with state-of-the-art methods designed for closed-set settings in unsupervised domain adaptation for event-based semantic segmentation.
\our demonstrates superior performance, showcasing its potential for real-world applications. The code is available at \href{https://github.com/ram95d/OVOSE}{https://github.com/ram95d/OVOSE}. %, particularly in dynamic environments.


\end{abstract}

\section{Introduction}
\label{sec:intro}

Event cameras, known for their exceptional temporal resolution, low latency, and motion blur resistance, have transformed various deep learning applications \cite{gallego2020event}.
Use cases include autonomous driving \cite{chen2020event}, object recognition \cite{Cho_2023_ICCV}, and semantic segmentation \cite{Binas2017DDD17ED}. Their outstanding performance in challenging conditions makes event cameras optimal for capturing reliable visual data \cite{messikommer2022hdrev}. Despite their success, integrating event cameras into existing computer vision models is challenging. Their unique data format, featuring asynchronous event streams without traditional image frames \cite{sun2022ess}, necessitates a reassessment of established techniques. While traditional image-based semantic segmentation has made notable progress \cite{Liang2022OpenVocabularySS}, event cameras, being less prevalent in real-world scenarios, suffer from a scarcity of both raw and labelled data. This fact raises two intertwined issues: the impossibility of collecting internet-scale event datasets and, as a consequence, the difficulty in training data-intensive deep learning techniques. This challenge obstructs effective semantic segmentation model training, especially in scenarios lacking established benchmarks or methodologies.

\begin{figure}[!t]
    \centering
    \includegraphics[width=\columnwidth]{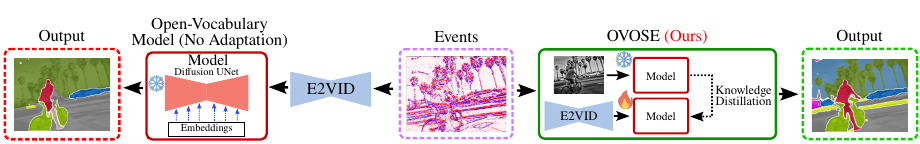}
    \caption{Output of a regular RGB foundation model for semantic segmentation and \our in event-based data. OVOSE accurately segments person, trees, and (the sky).}
    \label{fig:teaser}
\end{figure}
Recently, powerful foundation models have demonstrated their effectiveness in segmenting conventional images \cite{Liang2022OpenVocabularySS}.
Some of these models extend the closed-set capabilities of classical semantic segmentation models \cite{liang2018deeplab,liang2018v3}, performing well in open-vocabulary settings where the set of classes in training and testing are disjoint.
However, applying these powerful open-vocabulary models directly to event-based images is challenging, and retraining them is impractical due to the scarcity of annotated event-based data \cite{Jian_2023_ICCV}.

The alternative approach of converting events into images using E2VID {\cite{Rebecq19pami}}, and pairing it with an open-vocabulary model does not perform satisfactorily as illustrated in Fig.~\ref{fig:teaser}.
Despite the fact that E2VID is intended to reduce the domain gap between images and events, there still exists a difference between real grayscale and reconstructed-event images.
In this paper, we introduce \our, the first Open-Vocabulary Semantic Segmentation algorithm tailored for event-based data.
\our operates as a two-branch network, with one branch dedicated to grayscale images and the other to event data.
Each branch incorporates a copy of an image foundation model, with \our adapting the event branch for optimal performance in event-based data.
Our algorithm integrates text-to-image diffusion \cite{ramesh2022hierarchical} and a mask generator.
Using a CLIP-style image encoder and MLP, we derive embeddings for conditioning the text-to-image diffusion UNet.
We use UNet's features as input to the mask generator for mask generation.
We categorize the mask generator's outputs using a frozen CLIP-style text encoder for open-vocabulary segmentation.
To enhance model performance, we distill knowledge from the image branch to the events branch, using an E2VID model \cite{Rebecq19pami} for translating events to reconstructed images.
\our, characterized by its simplicity and effectiveness, outperforms existing models in event-based semantic segmentation, including Unsupervised Domain Adaptation (UDA) methods, demonstrating its effectiveness in addressing the open-vocabulary segmentation problem in event-based data.

Our main contributions can be summarized as follows:
\begin{itemize}
     \item To the best of our knowledge, we present the first open-vocabulary semantic segmentation approach tailored explicitly for event-based data.
    \item We introduce \our that distills knowledge to transfer semantic insights from a foundation model trained on regular images to enhance open-vocabulary segmentation performance for event-based data.
    \item To mitigate the effects of sub-optimal reconstructions, we investigate various mask reweighting strategies and introduced a novel dissimilarity network.
    This network recalibrates the mask loss by leveraging the differences between reconstructed and original images, enabling precise fine-tuning of the segmentation model and thus producing robust and accurate predictions.
    \item We perform extensive evaluations in open-vocabulary semantic segmentation for three event datasets.
    \our readily outperforms existing closed-set semantic segmentation methods and straightforward adaptations of open-vocabulary models.
    A set of ablation studies validates the key components of our algorithm. 
\end{itemize}
\section{Related Work}

\noindent\textbf{Event Camera Semantic Segmentation.}
Alonso et al. \cite{alonso2019ev} introduced event camera semantic segmentation an Xception-type network for the DDD17 dataset but suffered from limitations in the quality of generated labels. Gehrig et al. \cite{gehrig2020video} improved performance substantially by utilizing a synthetic event dataset converted from videos. Wang et al. \cite{wang2021dual} explored knowledge distillation and transfer learning between images and events, though relying on labeled datasets. Messikommer et al. \cite{messikommer2022bridging} proposed a method aligning image and event embeddings but faced challenges with hallucinations \cite{sun2022ess}. Sun et al. \cite{sun2022ess} addressed domain gap reduction between events and images, yet required real event-based data. Yang et al. \cite{Yang_2023_ICCV} presented a self-supervised learning framework, but still relied on labeled datasets for fine-tuning. While existing methods are designed for closed-set semantic segmentation with limited known classes, our approach is an open-vocabulary segmentation method tailored for event-based cameras, trained solely on synthetic unlabeled datasets.

\noindent\textbf{Open-Vocabulary Semantic Segmentation.}
Recent approaches in open-vocabulary semantic segmentation for regular images have centered on embedding spaces linking image pixels to class descriptors \cite{bucher2019zero,xian2019semantic}. Some methods leverage CLIP \cite{radford2021learning} for text and image embeddings \cite{li2022languagedriven}, while others combine CLIP with Vision Transformer \cite{dosovitskiy2020image}. OpenSeed is introduced \cite{zhang2023simple} for joint segmentation and detection tasks. ODISE \cite{xu2023open} uniquely merges pre-trained text-image diffusion and discriminative models \cite{rombach2022high}, excelling in open-vocabulary panoptic segmentation. However, these methods are designed for image data and don't directly translate to event cameras due to the fundamentally different data representation (continuous event streams vs. static images).
This work proposes a novel approach for open-vocabulary semantic segmentation on event cameras. We bridge the gap between image-based methods and event data by transferring knowledge from a powerful image foundation model to a new model specifically tailored for the event domain.

\subsection{Knowledge Distillation}
Prior knowledge distillation (KD) \cite{hinton2015distilling} methods focus on single modality, \ie, image using logits \cite{xu2020knowledge} or features \cite{park2019feed} or across different modalites \cite{Zhao2020KnowledgeAP} utilizing paired data.
KD is also applied to event cameras \cite{wang2021evdistill} to distill knowledge from the image-based teacher model to the event-based student model. 
However, they require labeled image datasets and unlabelled real events and frames for effective transfer in semantic segmentation.
Furthermore, \cite{wang2021dual} employs event-to-image transfer for semantic segmentation, but it gives poor performance \cite{sun2022ess} when applied for semantic segmentation.
Unlike these approaches, we employ a synthetic training dataset and distill knowledge from an image foundation model to a foundation model tailored for open vocabulary semantic segmentation in events.
The problem becomes complex as we strive to bridge the gap between events and images, and additionally tackle the synthetic-to-real gap.

\section{Open-vocabulary Segmentation in Events}

\begin{figure*}[!t]
    \centering
    \includegraphics[width=\textwidth]{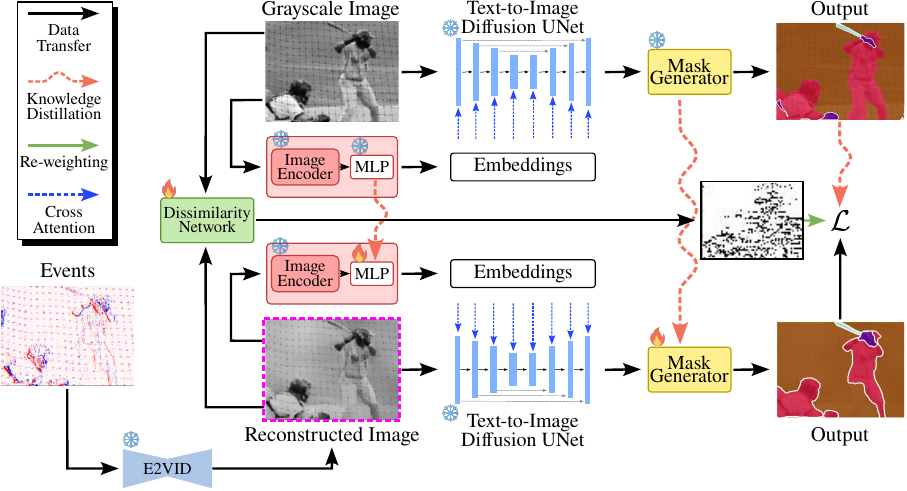}
    \caption{\textbf{Overview of \our pipeline}.
    Our algorithm comprises two components: the original grayscale image branch and the event-based branch. 
    Initially, events are transformed into a grayscale image using the E2VID model.
    Subsequently, both the original and reconstructed grayscale images undergo text embedding through an image encoder and an MLP. 
    The features from a frozen text-to-image diffusion UNet are then extracted for each tuple of image and text embedding.
    For each branch, a mask generator predicts class-agnostic binary masks and associated mask embedding features. 
    Categorization is achieved through a dot product between mask embedding features and text embeddings.
    Both branches are initialized with ODISE weights \cite{xu2023open}, and knowledge distillation occurs from the original image branch to the event-based branch during training.
    Original and reconstructed images are input into a dissimilarity network to weigh the distillation in the outputs.
    During the evaluation, only the event-based branch is utilized.}
    \label{fig:Pipeline}
\end{figure*}
\subsection{Preliminaries.}
\noindent \textbf{Event Representation}.
Each pixel in an event camera operates independently and reports brightness changes asynchronously, signaling only when the changes exceed a certain threshold. When a change is detected, an event is generated, capturing the pixel positions $(x_i, y_i)$, timestamp $(t_i)$, and polarity $(p_i)$, indicating whether a change involves an increase or decrease in brightness. Consequently, each event can be represented as $\mathbf{E} = [x_i, y_i, t_i, p_i]$. In this paper, we transform events into grid-like representations \cite{gehrig2019end}, such as voxel grid \cite{zhu2019unsupervised,sun2022ess} to facilitate further processing. We utilize a voxel grid representation of events as input to our model.

\noindent \textbf{Text-to-Image Diffusion UNet}.  The text-to-image diffusion model {\cite{ramesh2022hierarchical}} generates high-quality images from textual descriptions, utilizing the power of pre-trained encoders like CLIP {\cite{radford2021learning}} to encode text into embeddings. The process starts by adding Gaussian noise to images, and then the UNet architecture effectively reverses this noise, guided by cross-attention mechanisms that align the text embeddings with visual features, making the features semantically rich.  We incorporated the text-to-image diffusion UNet into our model to extract rich features that are relevant to the text. \\
\noindent \textbf{Problem statement}. Open vocabulary represents a generalization of the zero-shot task in semantic segmentation.
In this setting, a model predicts masks for unseen classes $\mathcal{C}^{unseen}$ by learning from labeled data of seen classes $\mathcal{C}^{seen}$.
The sets of seen and unseen classes are separate and do not overlap, \ie, $\mathcal{C}^{unseen} \cap \mathcal{C}^{seen} = \emptyset$.
The objective of this work is to train a model $F_\param$ with parameters $\param$ to predict the segmentation map of some stream of events.
To solve this, we have a unlabeled training set $\mathcal{X}_{train} = \{ \mathbf{E}_i^{(t)}, \mathbf{X}_i^{(t)} \}_{i=1}^{N_{t}}$, where $\mathbf{E}_i^{(t)}$ is the $i$th stream of events, $\mathbf{X}_i^{(t)} \in \mathbb{R}^{H \times W}$ is the $i$th original grayscale image with $H$ and $W$ the height and width of the image, and $N_t$ is the number of streams in the dataset.
For the training set, we additionally need the set of seen classes $\mathcal{C}^{seen}$.
We evaluate $F_\param$ in a testing set $\mathcal{X}_{test} = \{ \mathbf{E}_i^{(s)}, \mathbf{Y}_i^{(s)} \}_{i=1}^{N_{s}}$ where $\mathbf{Y}_i^{(s)} \in \{0,1\}^{H \times W \times \vert \mathcal{C}^{unseen} \vert}$.

\subsection{Overview of \our}

As shown in Figure \ref{fig:Pipeline} \our is divided into two sections: (i) the image branch that takes as input the original grayscale images $\mathbf{X}_i^{(t)} \in \trainset$, and (ii) the event branch that takes as input the stream of events $\ev_i^{(t)} \in \trainset$ or $\ev_i^{(s)} \in \testset$.
It is worth clarifying that during the evaluation we only use the event branch.
%we only use the event branch during evaluation.
During training, the whole image branch is frozen and only used to distill knowledge to the event branch.
For the event branch, we use the pre-trained E2VID model with parameters $\param_{e}$ \cite{Rebecq19pami} to transform some stream of events $\ev$ into a reconstructed image $\hat{\mathbf{X}}$, thus $F_{\param_e}(\ev) = \hat{\mathbf{X}}$. E2VID introduces a novel approach by leveraging a convolutional recurrent neural network architecture to process event camera data's sparse and asynchronous nature, producing high temporal resolution images.
Taking as input $\mathbf{X}$ or $\hat{\mathbf{X}}$, the forward pass of each branch is identical, so we explain only one of these in the following.

We first employ a frozen image encoder of type CLIP $\mathcal{V}(\cdot)$ \cite{radford2021learning} to encode $\mathbf{X}$ or $\hat{\mathbf{X}}$ into an embedding space.
Subsequently, a learnable MLP is used to project the image embedding into implicit text embeddings.
We use $\mathbf{X}$ or $\hat{\mathbf{X}}$ along with the implicit text embeddings as input to a text-to-image diffusion model \cite{Liang2022OpenVocabularySS} for feature extraction.
More precisely, we employ a UNet architecture to do the denoising process.
Formally for the image branch, we have:
\begin{equation}
    \mathbf{f} = F_{\param_U}(\mathbf{X},\MLP(\mathcal{V}(\mathbf{X}))),
\end{equation}
where $\mathbf{f}$ is the feature vector from the diffusion network in the image branch, and $\param_U$ is the parameters of the UNet model.
We feed this feature vector $\mathbf{f}$ as input to a Mask2Former model \cite{cheng2022masked} to produce class-agnostic binary masks with their corresponding mask embeddings.
For categorization, we use a text encoder of type CLIP $\mathcal{T}(\cdot)$ to embed the categories in $\mathcal{C}^{seen}$.
We thus perform a dot product between text and mask embeddings to categorize each mask.

We use the estimated segmentation map of the image branch as the ground truth of the event branch by computing a loss function between both outputs.
However, as we kept E2VID frozen, under poor reconstructions, we weighed this loss function using the output of a dissimilarity network to give more emphasis to the regions where E2VID reconstructs well.
We further perform knowledge distillation from the Mask2Former in the image to the Mask2Former in the event branch, and similarly for the MLP networks.
The parameters of the two branches are initialized with the weights of the ODISE model \cite{xu2023open}. 

\subsection{Distilling Image Embeddings}
To address the difference in image embeddings between the output of the MLP for the grayscale image and the corresponding output for the synthetic (reconstructed) image, we implement knowledge distillation. This involves transferring knowledge from the image embeddings of the original image to those of the reconstructed image in the event branch. 
To do so, we introduce the minimization of the Frobenius norm $ \left( \| \cdot \|_F \right)$  of the matrix of differences between real and reconstructed images encoded by the trainable MLP:
\begin{equation}
\mathcal{L}_{t} = \| \MLP_{\mathbf{X}}(\mathcal{V}(\mathbf{X})) - \MLP_{\ev}(\mathcal{V}(\hat{\mathbf{X}}))\|_F,
\label{eqn:mlp_distillation}
\end{equation}
where $\MLP_{\mathbf{X}}(\cdot)$ is the frozen MLP of the image branch and $\MLP_{\ev}(\cdot)$ is the MLP of the event branch. In other words, we leverage the information encoded in the image embeddings of the original image to guide the learning of the image embeddings for the reconstructed image.
The Frobenius norm serves as a metric to quantify the dissimilarity between these embeddings, enabling the model to refine its representation of image information and enhance the consistency between the two image modalities.

\subsection{Feature Distillation}

To provide further guidance from the image branch to the event one, we minimize the Frobenius norm of the matrix differences of the outputs of each layer in the transformer decoder of Mask2Former between the original image and reconstructed image:
\begin{equation}
\mathcal{L}_{f} = \frac{1}{L} \sum_{i=1}^L \| \mathbf{D}_i - \mathbf{\widehat{D}}_i \|_F \,,
\end{equation}

where $L$ represents the total number of decoder layers, and $\mathbf{D}$ and $\mathbf{\widehat{D}}$ are the output matrices of each layer in the image and events branch, respectively.

\subsection{Mask Re-weighting}

\begin{figure}[t]
    \centering
    \includegraphics[width=\columnwidth]{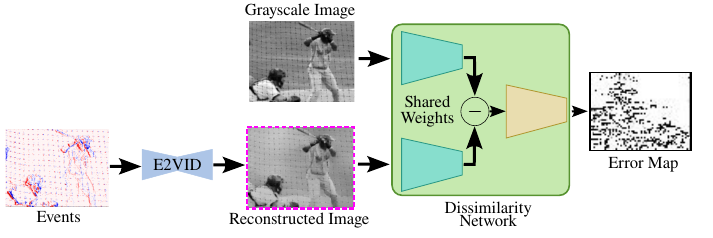}
    \caption{Dissimilarity network takes the grayscale and reconstructed images as input, and it outputs an error map to reweight the mask loss. E2VID is unable to reconstruct the stripes and hence considered a high error area by the dissimilarity network.}
    \label{fig:dissimilarity_network}
\end{figure}
\begin{figure}[t]
    \centering
    \includegraphics[width=\columnwidth]{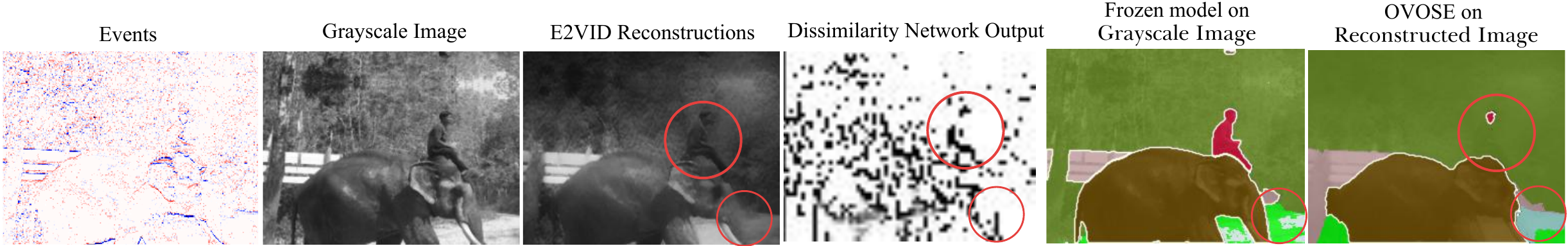}
    \caption{The impact of reweighting the mask loss, influenced by the dissimilarity between the grayscale and reconstructed images. Poorly reconstructed areas such as the person and the elephant's trunk lead to their exclusion in the reweighting process.}
    \label{fig:Diss}
\end{figure}

While maintaining E2VID frozen, poor reconstructions may occur, prompting the imposition of classification on inadequately reconstructed regions. However, this approach risks compromising model performance where reconstructions are accurate. To address this, we introduce a dissimilarity network to discern differences between the grayscale image and its reconstructed counterpart from events. Illustrated in Figure \ref{fig:dissimilarity_network}, this network comprises two convolutional layers. The first layer shares weights for grayscale and reconstructed images, followed by a rectified linear unit ($\relu$) activation, while the second layer is followed by a sigmoid activation function $\sigma(\cdot)$. The squared error between the grayscale and reconstructed images' outputs feeds into the second convolutional layer, generating a reweighting map. As shown in Figure~\ref{fig:dissimilarity_network} that a notable discrepancy exists between the grayscale image and the reconstructed image, particularly concerning the stripes on the shirt in this example.
Consequently, this discrepancy in the error map indicates reduced importance attributed to that specific area.
Mathematically, this process can be expressed as:
\begin{equation}
    \begin{aligned}
        &\mathbf{M} = \sigma(\conv_{2}(\relu(\conv_{1}(\mathbf{X})) - \relu(\conv_{1}(\hat{\mathbf{X}})))^2),
    \end{aligned}
\end{equation}

where $\mathbf{X}$ is the grayscale original image and $\hat{\mathbf{X}}$ is the reconstructed image from E2VID.
We use $\mathbf{M}$ to re-weight our distillation loss at the level of segmentation maps $\mathcal{L}_{m}$. 
This loss for the $i$th stream is given by:
\begin{equation}
    \mathcal{L}_{m} = \mathbf{M} \odot \mathcal{L}_{CE} \left( \mathbf{Y}_i, \hat{\mathbf{Y}}_i \right),
\end{equation}
where $\mathbf{Y}_i \in \mathbb{R}^{H \times W \times \vert \mathcal{C}^{seen} \vert}$ is the output segmentation map of the image branch, $\hat{\mathbf{Y}}_i \in \mathbb{R}^{H \times W \times \vert \mathcal{C}^{seen} \vert}$ is the output segmentation map of the event branch, $\mathcal{L}_{CE}$ is the cross-entropy loss, and $\odot$ is the point-wise product between matrices.  We illustrate a sample output of the dissimilarity network in Figure ~\ref{fig:Diss}. It can be seen that \our successfully ignores the areas where the error is high, for example, the person and the elephant's trunk.

\subsection{Category Label Supervision}
As we have access to category labels in the training set, we follow \cite{xu2023open} to compute a category label loss $\mathcal{L}_c $.
To this end, we compute the probability of a mask belonging to one of the training categories using a cross-entropy loss with a learnable temperature parameter as in \cite{xu2023open}.

The total loss of \our is given by:
\begin{equation} \label{eq:6}
\mathcal{L}_{final} = \mathcal{L}_t + \mathcal{L}_f + \lambda_{m}\mathcal{L}_m + \lambda_{c}\mathcal{L}_c 
\end{equation}
where $\lambda_{m}=5.0$ and $\lambda_{c}=2.0$ are regularization parameters.
%-------------------------------------------------------------------------
\section{Experiments and Results}

In this section, we present an overview of the baseline methods for comparison, details on the training and test data, and a comprehensive analysis of both quantitative and qualitative results.
Subsequently, we delve into ablation studies to validate the components of \our.

\subsection{Experimental Framework}
\noindent {\textbf{Baseline methods}.}
As there is no open-vocabulary semantic segmentation method for events, we benchmark \our against leading UDA methods for semantic segmentation in event-based data.
However, this comparison is unfair with \our since UDA methods: (i) know the set of unseen classes, (ii) have access to one or multiple labeled source datasets, and (iii) have access to the unlabeled testing dataset for adaptation.
We also compare our algorithm with straightforward adaptations of open-vocabulary semantic segmentation methods in regular images.
This adaptation consists of reconstructing a grayscale image from the stream of events with the E2VID model (similar to our event branch) and using this as input to the open-vocabulary model.
More precisely, \our is compared with the UDA methods E2VID \cite{Rebecq19pami}, EV-transfer \cite{messikommer2022bridging}, VID2E \cite{gehrig2020video} and ESS \cite{sun2022ess}.
For the open-vocabulary methods, we compare our algorithm against E2VID+OpenSeed \cite{zhang2023simple} and E2VID+ODISE \cite{xu2023open}.
We use the mean Intersection over Union (mIoU) and pixel accuracy metrics for the quantitative evaluations.

\noindent {\textbf{Training data}.}
As described in the problem statement, \our requires a training dataset $\mathcal{X}_{train}$ where we have the stream of events and their corresponding grayscale images.
To address this need, we leverage synthetic training data as introduced by \cite{Rebecq19pami}.
This synthetic dataset was generated using the event simulator ESIM \cite{Rebecq2018ESIMAO}, which simulated MS-COCO images \cite{lin2014microsoft}.
The dataset comprises $1,000$ sequences, each spanning $2$ seconds, with grayscale images and events of dimensions $240\times180$.
For our training purposes, we resize the images and events to dimensions $256\times192$, ensuring compatibility and optimization for our network.
Following \cite{xu2023open}, we use MS-COCO classes for category label supervision.

\noindent {\textbf{Evaluation datasets}.}
We evaluate the open-vocabulary performance of \our on two popular event camera-based self-driving datasets and TimeLens++:
\begin{itemize}
    \item \textit{DAVIS Driving Dataset} (DDD17) is a dataset for semantic segmentation in autonomous driving. 
    Alonso et al. \cite{alonso2019ev} pre-trained an Xception network to generate semantic pseudo-labels which were consolidated into six classes: flat (road and pavement), background (construction and sky), object (pole, pole group, traffic light, traffic sign), vegetation, human, and vehicle.
    \item \textit{DSEC-Semantic} DSEC \cite{Gehrig2021DSECAS} consists of high-resolution images $1440\times1080$, synchronized events $640\times480$, and semantic labels \cite{sun2022ess} are generated using a state-of-the-art semantic segmentation method. 
    The fine-grained labels for $19$ classes are consolidated into $11$ classes: background, building, fence, person, pole, road, sidewalk, vegetation, car, wall, and traffic sign.
    \item \textit{TimeLens++} We show qualitative results of applying directly \our in the Time lens++ dataset \cite{Tulyakov22CVPR}. Time Lens++ consists of high-resolution events of size $970\times625$.
\end{itemize}

\noindent {\textbf{Implementation details}.}
We keep the image branch frozen and fine-tune the MLP and Mask2Former of the event branch. 
We use convolutions with $3\times3$ kernel size with a stride of $2$ and padding $1$ for the dissimilarity network. 
We set the learning rate as $1\times10^{-5}$. 
We train \our using Adam optimizer with a batch size of $4$ on Nvidia Ampere GPU with 48GB of RAM.
We initialize the weights of text and image encoder from \cite{radford2021learning}, text-to-image diffusion UNet from \cite{ramesh2022hierarchical}, and MLP and Mask2former from \cite{xu2023open}.
For qualitative results we use the open source code and weights provided by \cite{sun2022ess} and \cite{xu2023open}. 
For OpenSeed \cite{zhang2023simple}, we use their provided open-source code and weights.

\subsection{Results}
\begin{figure*}[!t]
    \centering
    \includegraphics[width=\textwidth]{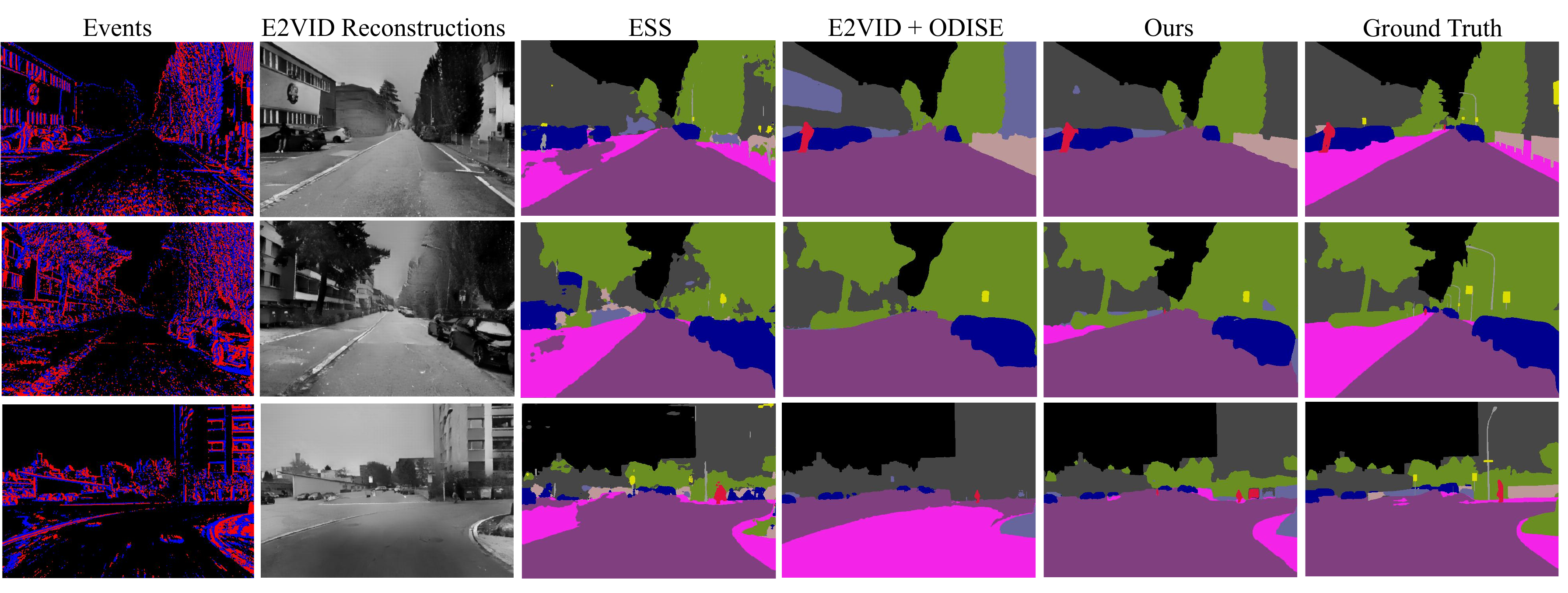}
    \caption{Qualitative samples from ESS in UDA closed-set, E2VID+ODISE, and \our in open vocabulary setting. As compared to ESS and E2VID+ODISE, \our produce accurate and less noisy predictions even though it is trained on a synthetic dataset.}
    \label{fig:dsec_qual}
\end{figure*}

\noindent {\textbf{Semantic segmentation on DSEC-Semantic}.}
Figure \ref{fig:dsec_qual} shows a qualitative comparison of \our against ESS and ODISE.
We obtain the qualitative results for the ESS in Figure \ref{fig:dsec_qual} by utilizing the official model provided by the authors.
Even though the ESS method is trained on real events in a closed-set setting, its overall semantic segmentation across the entire image appears noisy.
Notably, it fails to segment vehicles in several instances accurately.
For E2VID+ODISE in an open vocabulary setting, it misclassifies parts of the road and buildings.
In contrast, \our excels in recognizing traffic signs and delivers superior overall semantic segmentation across the entire image. 
This is particularly evident in its ability to discern intricate details and provide accurate segmentations, showcasing its robustness and adaptability in the driving scenario of DSEC.

Table \ref{tab:dsec} presents the comparison of \our against the baseline methods on the DSEC-Semantic dataset.
Even though the UDA methods are trained in the closed-set setting using real events and urban street datasets similar to DSEC, \our surpasses their performance by a significant margin. 
Notably, our model improves the state-of-the-art UDA method ESS \cite{sun2022ess} by a substantial $3.57\%$ in the mIoU metric.
Similarly, \our outperforms E2VID+ODISE by $4.83\%$ in the mIoU metric, showcasing the superior performance of \our for semantic segmentation in event-based data.
\renewcommand{\arraystretch}{1.2}
\begin{table*}[!htb]
  \caption{Results on DSEC Semantic in  UDA and open-vocabulary setting. \our not only outperforms all the UDA methods even though they are trained in closed-set settings but also translated open-vocabulary methods.}
  \centering
  \resizebox{\textwidth}{!}{
  \begin{tabular}{lcccccccc}
    \toprule
    \textbf{Method} & \textbf{Type} & \textbf{Problem Formulation} & \textbf{Training Data} & \textbf{Input} & \textbf{Acc (\%)} $\uparrow$ & \textbf{mIoU (\%)} $\uparrow$ \\
    \midrule
    EV-Transfer \cite{messikommer2022bridging} & UDA & Closed-Set &Cityscapes+DSEC& events+frames & $60.50$  & $23.20$ \\
    E2VID \cite{Rebecq19pami} & UDA & Closed-Set &Cityscapes+DSEC& events+frames & $76.67$  & $40.70$ \\
    ESS \cite{sun2022ess} & UDA & Closed-Set &Cityscapes+DSEC& events+frames & $84.04$  & $44.87$ \\
    \hdashline
    E2VID+OpenSeed \cite{zhang2023simple} & Translation & Open-Set& MS-COCO &events & $65.25$ & $32.82$\\
    E2VID+ODISE \cite{xu2023open} & Translation & Open-Set &  MS-COCO &events& $81.24$ & $43.61$\\
    OVOSE (Ours) & Distillation & Open-Set & EV-COCO & events & $\textbf{85.67}$ & $\textbf{48.44}$ \\
    \bottomrule
  \end{tabular}
  }
  \label{tab:dsec}
\end{table*}

\begin{figure*}[!t]
    \centering
    \includegraphics[width=0.99\textwidth]{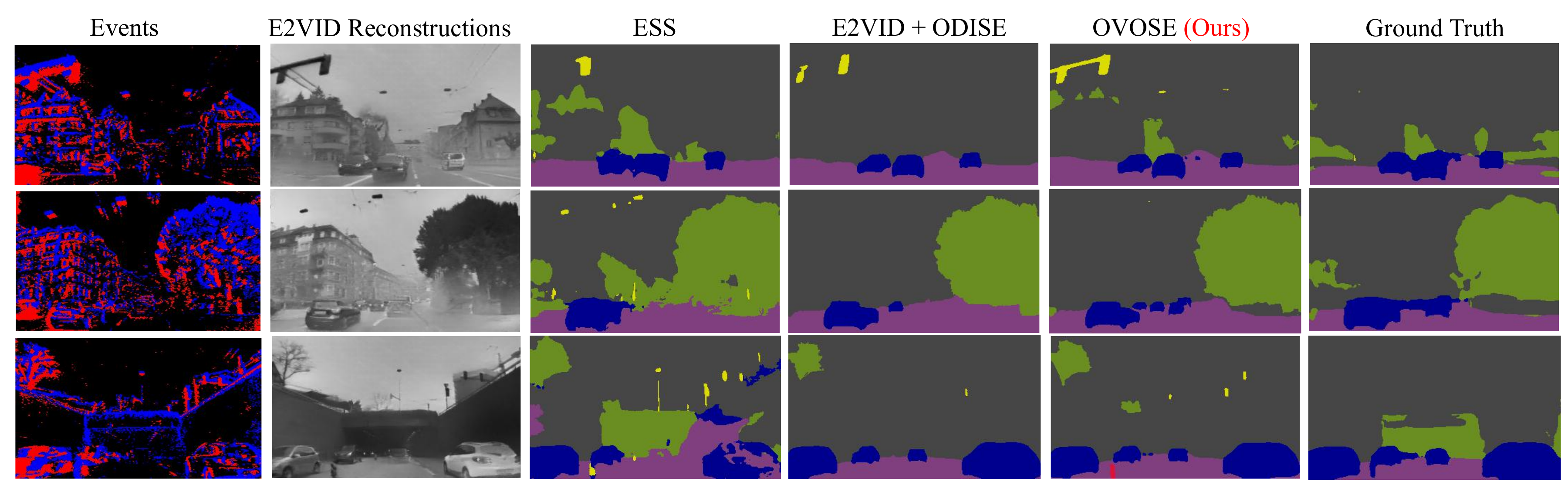}
    \caption{Qualitative samples from DDD17 in UDA closed-set, E2VID+ODISE, and \our in open vocabulary setting. \our does better overall predictions, especially vehicles, persons, vegetation, and construction.}
    \label{fig:ddd17_qual}
\end{figure*}

\noindent {\textbf{Semantic segmentation on DDD17}.}
Figure \ref{fig:ddd17_qual} presents a qualitative comparison between \our, ESS, and E2VID+ODISE.
ESS effectively segments poles and other event-specific objects that may not be available in the ground truth.
However, it struggles with inaccuracies and noise when segmenting cars.
In contrast, \our operating in an open vocabulary setting showcases the ability to segment traffic signs even when unlabeled in the ground truth images. 
Notably, \our excels in differentiating between classes, particularly with vehicles, and demonstrates more accurate and noise-resistant semantic segmentation compared to the ESS.
\our also outperforms E2VID+ODISE by providing clearer object recognition and segmentation. 
This comparison emphasizes the need for knowledge distillation and \our's capacity to deliver precise semantic segmentation even in scenarios with incomplete or noisy ground truth annotations.

\begin{figure*}[!t]
    \centering
    \includegraphics[width=0.99\textwidth]{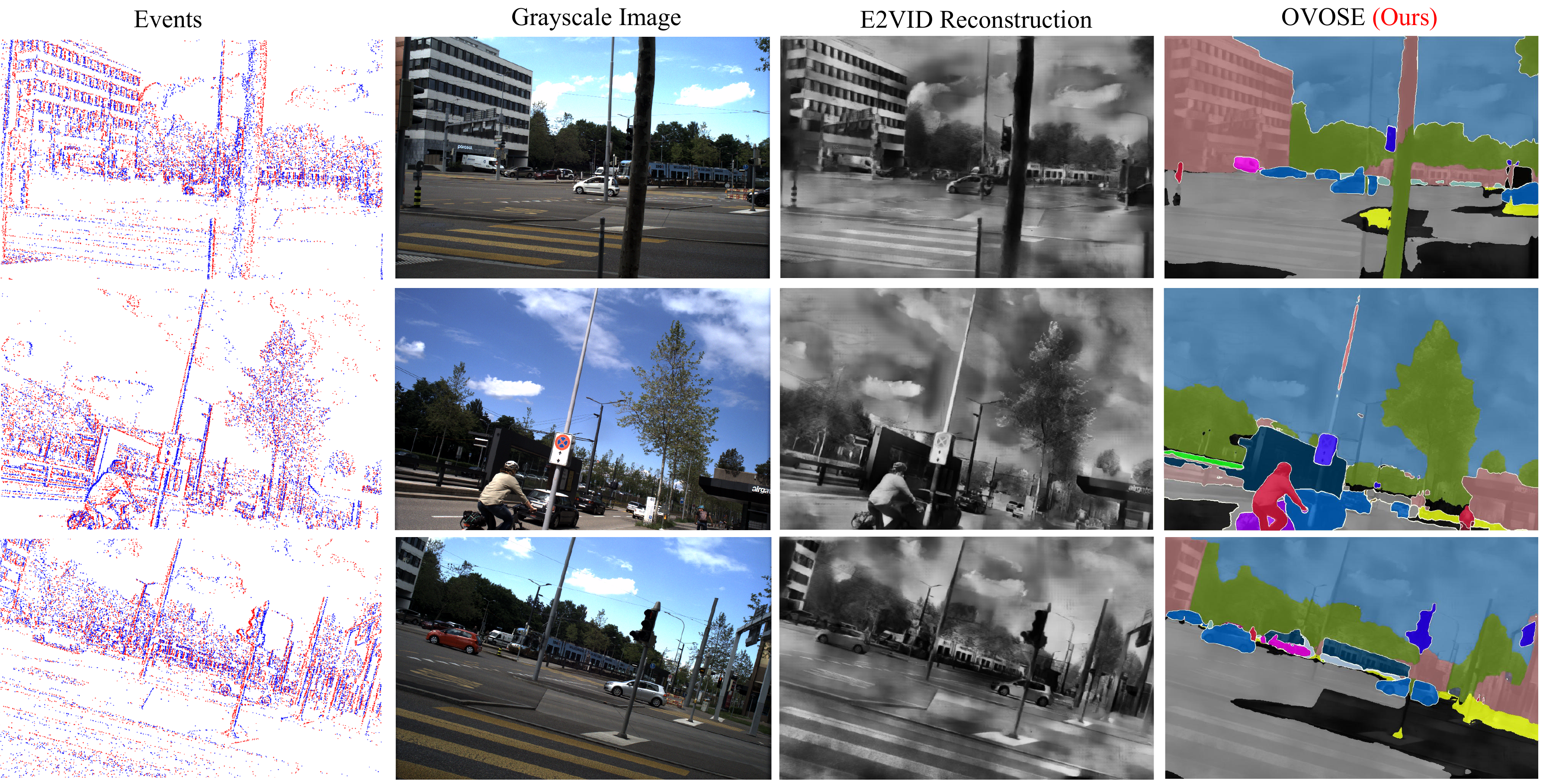}
    \caption{Open-vocabulary performance of \our on the Time Lens++ dataset \cite{Tulyakov22CVPR}.}
    \label{fig:tl_qual}
\end{figure*}

When it comes to quantitative evaluation on the DDD17 dataset, the presence of noisy ground truth labels, as highlighted by \cite{sun2022ess}, poses challenges. 
Additionally, the previous work by Ev-SegNet \cite{alonso2019ev} merged several labels, further complicating the evaluation process.
To address this, we consider the original classes pre-merge for predictions, subsequently merging them to facilitate a comparison with the ground truth.
Table \ref{tab:dd17} summarizes the quantitative comparison of \our with the baseline methods on the DDD17 dataset.
\our significantly improves the UDA state-of-the-art ESS method \cite{sun2022ess} by $0.93\%$ in the mIoU metric.
Notably, our algorithm outperforms E2VID+OpenSeed and E2VID+ODISE by large margins, underscoring the significance of knowledge distillation and mask reweighting in semantic segmentation for event cameras.
\begin{table*}[!t]
  \caption{Results on DDD17 in  UDA and open-vocabulary setting. \our outperforms all the UDA methods trained in closed-set settings but also translated open vocabulary methods}
  \centering
  \resizebox{\textwidth}{!}{
  \begin{tabular}{lccccccc}
    \toprule
    \textbf{Method} & \textbf{Type} & \textbf{Problem Formulation} & \textbf{Training Data} & \textbf{Input} & \textbf{Acc (\%)} $\uparrow$ & \textbf{mIoU (\%)} $\uparrow$ \\
    \midrule
    EV-Transfer \cite{messikommer2022bridging} & UDA & Closed-Set & Cityscapes+DDD17& events+frames & $47.37$  & $14.91$ \\
    E2VID \cite{Rebecq19pami} & UDA & Closed-Set & Cityscapes+DDD17 & events+frames & $83.24$  & $44.77$ \\
    VID2E \cite{gehrig2020video} & UDA & Closed-Set & Cityscapes+DDD17& events+frames & $85.93$  & $45.48$ \\
    ESS \cite{sun2022ess} & UDA & Closed-Set & Cityscapes+DDD17& events+frames & $87.86$  & $52.46$ \\
    \hdashline
    E2VID+OpenSeed \cite{zhang2023simple} & Translation & Open-Set & MS-COCO &events& $33.25$ & $17.95$ \\
    E2VID+ODISE \cite{xu2023open} & Translation & Open-Set & MS-COCO &events& $84.63$ & $48.12$\\
    OVOSE (Ours) & Distillation & Open-Set & EV-COCO & events & $\textbf{88.84}$ & $\textbf{53.39}$\\

    \bottomrule
  \end{tabular}
  }
  \label{tab:dd17}
\end{table*}
\noindent {\textbf{Qualitative Results on the Time Lens++ Dataset}}. We use the Time Lens++ \cite{Tulyakov22CVPR} dataset to evaluate the open-vocabulary capabilities of \our since no open-vocabulary event dataset is available.
Some qualitative results are illustrated in Figure~\ref{fig:tl_qual}.
We see in Figure~\ref{fig:tl_qual} that \our performs high-quality segmentation of various classes such as trees, traffic lights, people, cars, buildings, sidewalk, and roads which demonstrates \our preserved open vocabulary capabilities. Furthermore, \our can recognize the bike and train even though the reconstruction from E2VID is very noisy and far from optimal.
\subsection{Ablation Studies}
\noindent \textbf{Mask Reweighting.}
We conduct an ablation study to validate the reweighting schemes and knowledge distillation techniques in \our.
To that end, We evaluate \our without training; this is equivalent to performing E2VID+ODISE. Then, we analyzed the impact of reweighting by using only the distillation pipeline, with reweighting based upon cosine similarity (CS) on the original and reconstructed image, by using the squared differences of the output of Stable Diffusion (SD) from the image and event branches and finally with our Dissimilarity Network (DN) corresponding to \our.
\begin{table}[t]
  \caption{Ablation of distillation strategies of \our on DSEC Semantic dataset.}
  \centering
  \renewcommand{\arraystretch}{0.9} % Reduce row spacing
  \begin{tabular}{ccccc}
    \toprule
    \textbf{Method} & \textbf{Acc (\%)} $\uparrow$ & \textbf{mIoU (\%)} $\uparrow$ \\
    \midrule
    Baseline & $81.2$ & $43.61$\\
    Distillation & $85.0$ & $47.01$\\
    Distillation+Reweight CS & $85.1$ & $48.01$\\
    Distillation+Reweight SD & $85.1$ & $47.32$\\
    Distillation+Reweight DN & $\textbf{85.6}$ & $\textbf{48.44}$\\
    \bottomrule
  \end{tabular}
  \label{tab:abl}
\end{table}

\begin{table}[t]
  \centering
  \caption{Ablation studies on DSEC-Semantic dataset. MLP's is MLP layer of OVOSE and MG is the Mask Generator}.
  \renewcommand{\arraystretch}{0.9} % Reduce row spacing
  \begin{tabular}{cccccc}
    \toprule
    \textbf{Ablation} & \multicolumn{2}{c}{\textbf{Parameters}}  & \textbf{Acc (\%)} $\uparrow$ & \textbf{mIoU (\%)} $\uparrow$ \\
    \midrule
    \textbf{} & \textbf{\underline{$\boldsymbol{\lambda_c}$}} & \textbf{\underline{$\boldsymbol{\lambda_m}$}}  & & \\
    % \midrule
    \multirow{3}{*}{Loss} & $5.0$ & $5.0$ & $84.47$ & $46.47$ \\
                            & $5.0$ & $2.0$ & $84.70$ & $46.44$ \\
                            & $2.0$ & $5.0$ & $\textbf{85.67}$ & $\textbf{48.44}$ \\
    \midrule
    \textbf{} & \textbf{\underline{MLP's}} & \textbf{\underline{MG}}  &  &  \\
    \multirow{3}{*}{Finetuning} & \checkmark &  \ding{55} & $82.46$ & $44.34$ \\
                                & \ding{55} & \checkmark & $83.19$ & $44.68$ \\
                                & \checkmark & \checkmark & $\textbf{85.67}$ & $\textbf{48.44}$ \\
    \midrule
    \textbf{} & \textbf{\underline{FireNet \cite{Scheerlinck20wacv}}} & \textbf{\underline{E2VID \cite{Rebecq19pami}}}  &  &  \\
    \multirow{2}{*}{Image-Reconstructor} & \checkmark & \ding{55} & $82.85$ & $43.94$ \\
                                         & \ding{55} & \checkmark & $\textbf{85.67}$ & $\textbf{48.44}$ \\
    \midrule
    \textbf{} & \textbf{\underline{DSEC Classes}} & \textbf{\underline{Ours}}  &  &  \\
    \multirow{2}{*}{Text Prompts} & \checkmark & \ding{55} & $84.04$ & $46.50$ \\
                                         & \ding{55} & \checkmark & $\textbf{85.67}$ & $\textbf{48.44}$ \\
    \bottomrule
  \end{tabular}
  \label{tab:comprehensive}
\end{table}
Table \ref{tab:abl} shows the results of the ablation study.
Distillation resulted in a notable increase in accuracy by $3.8\%$ and mIoU by $3.4\%$, showcasing the effectiveness of this approach.
Moreover, the introduction of the dissimilarity network for reweighting the loss function gives an additional improvement of $1.43\%$ in the mIoU metric.
\noindent \noindent \textbf{Loss Parameters.}
We study the impact of varying $\lambda_c$ and $\lambda_m$ in Equation {~\ref{eq:6}} on model performance. As shown in Table ~\ref{tab:comprehensive}, the optimal combination of $\lambda_c = 2.0$ and $\lambda_m = 5.0$ achieved the highest accuracy $85.67\%$ and mean Intersection over Union (mIoU) $48.44\%$, indicating that a lower weight on caption loss and a higher weight on mask loss enhance performance.

\noindent \noindent \textbf{Fine-tuning Ablation.}
We explore the effects of fine-tuning the MLP layers and the Mask Generator (MG), individually and together. From Table ~\ref{tab:comprehensive}, we observe that fine-tuning only the MLP resulted in $82.46\%$ accuracy and $44.34\%$ mIoU, while fine-tuning only MG slightly improved performance to $83.19\%$ accuracy and $44.68\%$ mIoU. However, simultaneous fine-tuning of both MLP's and MG yielded the best results with $85.67\%$ accuracy and $48.44\%$ mIoU, underscoring the necessity of fine-tuning both components together.

\noindent \noindent \textbf{Image Reconstructor Ablation.}
We replaced E2VID {\cite{Rebecq19pami}} with FireNet \cite{Scheerlinck20wacv} and reported results in Table ~\ref{tab:comprehensive}. E2VID significantly outperformed FireNet, achieving $85.67\%$ accuracy and $48.44\%$ mIoU which indicates that E2VID's recurrent neural network handles temporal information well and outputs higher quality reconstructions for our downstream task.

\noindent \noindent \textbf{Ablation on Text prompts.}
We further assess the effectiveness of two text prompt configurations: directly using the DSEC-Class name and our text prompts shown in Table 1 and Table 2 in the supplementary document. Results in Table ~\ref{tab:comprehensive} show that DSEC-Classes configuration resulted in $84.04\%$ accuracy and $46.50\%$ mIoU. In contrast, our text prompts configuration achieved superior performance with $85.67\%$ accuracy and $48.44\%$ mIoU, demonstrating that the detailed prompts enhance the model's performance more effectively.
\section{Conclusion}

In this work, we introduced \our, the first open-vocabulary semantic segmentation algorithm designed for event-based data.
Comprising grayscale image and event branches, each equipped with a pre-trained foundation model, our approach leverages synthetic data for Knowledge Distillation from regular images to enhance semantic segmentation in events.
\our employs distillation at multiple stages of the foundation model, enhancing its effectiveness with a mask reweighting strategy through a dissimilarity network.
We evaluate \our in the DDD17 and DSEC-Semantic datasets and compare it against with existing methods in UDA close-set semantic segmentation and foundation models adapted to the event domain with E2VID.
Our algorithm outperforms all these models, offering a promising avenue for research in open-vocabulary semantic segmentation tailored for event cameras.
\section*{Acknowledgments}
We acknowledge financial support from the PNRR MUR project PE0000013-FAIR and from the Sapienza grant RG123188B3EF6A80 (CENTS).
\bibliographystyle{splncs04}
\bibliography{main}
% Include the supplement without extra blank pages
\ifarXiv
    

\section*{Supplementary material}

# Supplementary

Muhammad Rameez Ur Rahman[1], Jhony H. Giraldo[2], Indro Spinelli[1], Stéphane Lathuilière[2], and Fabio Galasso[1]

[1] Sapienza University of Rome, Italy
[2] LTCI, Télécom Paris, Institut Polytechnique de Paris, France

In this supplementary document, we provide details about the text prompts used in our model, fully supervised results, and comprehensive ablation studies of our method.

## 1 Text Prompts

Tables 1 and 2 present the textual prompts used to create CLIP text embeddings for two datasets: DSEC-Semantic and DDD17. Table 1 outlines the classes and corresponding text prompts utilized for the DSEC-Semantic dataset. The prompts are designed to cover a wide range of visual entities and contexts to enhance the model's understanding. Each class is associated with various descriptive terms to ensure comprehensive semantic coverage. For instance, the "person" class includes prompts like "man," "women," "person," "pedestrian," and "children," which cater to different possible appearances and contexts of individuals in the dataset. Similarly, Table 2 lists the classes and their corresponding text prompts used in the DDD17 dataset.

**Table 1.** The textual prompts utilized in the DSEC-Semantic dataset [3] for creating the CLIP text embeddings.

| # | Classes and the corresponding Text Prompts |
|---|---|
| 0 | **background**: sky |
| 1 | **building**: building, skyscraper, house, |
| 2 | **fence**: fence |
| 3 | **person**: man, women, person, pedestrian, children |
| 4 | **pole**: pole |
| 5 | **road**: road, highway, expressway |
| 6 | **sidewalk**: sidewalk, track, footpath, foot pavement pedestrian zone |
| 7 | **vegetation**: vegetation, tree, trees, palm tree, bushes, plants, trees, lawn |
| 8 | **car**: car, bicycle, bike, truck, locomotive, bus, motorcycle |
| 9 | **wall**: wall, stone wall |
| 10 | **traffic-sign**: traffic sign, stop sign, traffic light |

**Table 2.** The textual prompts utilized in the DDD17 dataset [1] for creating the CLIP text embeddings.

| # | Classes and the corresponding Text Prompts |
|---|---|
| 0 | **flat**: road, highway, pavement, street, lane track |
| 1 | **background**: building, sky, house, wall, fence |
| 2 | **object**: pole, light pole, street light, traffic light, traffic signal, traffic sign |
| 3 | **vegetation**: vegetation, tree, grass, meadow, plants |
| 4 | **human**: human, person, rider, people |
| 5 | **vehicle**: vehicle, car, truck, bus, motorcycle, bicycle, train, van, locomotive |

## 2  DSEC supervised results

We compare OVOSE against the previous methods EV-SegNet [1], ESS [3] and HMNET [2] in a fully supervised setting. To that end, we train OVOSE in a supervised closed-set setting on the DSEC-Semantic dataset [3]. We report the results in Table 3 and it can be seen that OVOSE outperforms ESS by a large margin of 3.62% mIoU in Table 3.

**Table 3.** Comparison of OVOSE with state-of-the-art methods on DSEC Semantic dataset in a supervised setting. OVOSE outperforms state-of-the-art by 3.62% mIoU

| Method | Acc (%) ↑ | mIoU (%) ↑ |
|---|---|---|
| EV-SegNet [1] | 88.61 | 51.76 |
| ESS [3] | 89.25 | 51.57 |
| ESS [3] | 89.37 | 53.29 |
| HMNET-B3 [2] | 88.70 | 51.20 |
| HMNET-L3 [2] | **89.90** | 55.00 |
| OVOSE (ours) | 89.45 | **58.62** |

## References

1. Alonso, I., Murillo, A.C.: Ev-segnet: Semantic segmentation for event-based cameras. In: 2019 IEEE/CVF Conference on Computer Vision and Pattern Recognition Workshops (CVPRW). pp. 1624–1633 (2019)
2. Hamaguchi, R., Furukawa, Y., Onishi, M., Sakurada, K.: Hierarchical neural memory network for low latency event processing. In: Proceedings of the IEEE/CVF Conference on Computer Vision and Pattern Recognition. pp. 22867–22876 (2023)
3. Sun, Z., Messikommer, N., Gehrig, D., Scaramuzza, D.: Ess: Learning event-based semantic segmentation from still images. In: Avidan, S., Brostow, G., Cissé, M., Farinella, G.M., Hassner, T. (eds.) Computer Vision – ECCV 2022. pp. 341–357. Springer Nature Switzerland, Cham (2022)


\fi
% \ifarXiv
%     \foreach \x in {1,...,\numbersupplementpages}
%     {
%         % \clearpage
%         \includepdf[pages={\x,{}}]{\supplementfilename}
%     }
% \fi



% LNCS Homepage, \url{http://www.springer.com/lncs}. Last accessed 4
% \end{thebibliography}
\end{document}